\documentclass[10pt,twocolumn,letterpaper]{article}

\usepackage{cvpr}              

\usepackage{graphicx}
\usepackage{amsmath}
\usepackage{amssymb}
\usepackage{booktabs}
\usepackage{bbm}

\usepackage{bbding}
\usepackage{makecell}
\usepackage{multirow}
\usepackage{xcolor} 

\newcommand{\customfootnotetext}[2]{{
  \renewcommand{\thefootnote}{#1}
  \footnotetext[0]{#2}}}

\definecolor{linkc}{rgb}{0, 0.44, 0.74}
\usepackage[pagebackref,breaklinks,colorlinks, urlcolor=linkc]{hyperref}

\usepackage[capitalize]{cleveref}
\crefname{section}{Sec.}{Secs.}
\Crefname{section}{Section}{Sections}
\Crefname{table}{Table}{Tables}
\crefname{table}{Tab.}{Tabs.}

\def\cvprPaperID{1438} 
\def\confName{CVPR}
\def\confYear{2023}

\begin{document}

\title{ARKitTrack: A New Diverse Dataset for Tracking Using Mobile RGB-D Data}

\author{Haojie Zhao$^{1}$\textsuperscript{$\dag$} \quad Junsong Chen$^{1}$\textsuperscript{$\dag$} \quad Lijun Wang$^{1}$\textsuperscript{*} \quad Huchuan Lu$^{1,2}$\\
$^{1}$Dalian University of Technology, China \quad $^{2}$Peng Cheng Laboratory, China\\
{\tt\small \{haojie\_zhao,jschen\}@mail.dlut.edu.cn \quad \{ljwang,lhchuan\}@dlut.edu.cn}
}

\twocolumn[{%

\maketitle
\begin{center}
    \centering
    \captionsetup{type=figure}
    \includegraphics[width=\textwidth]{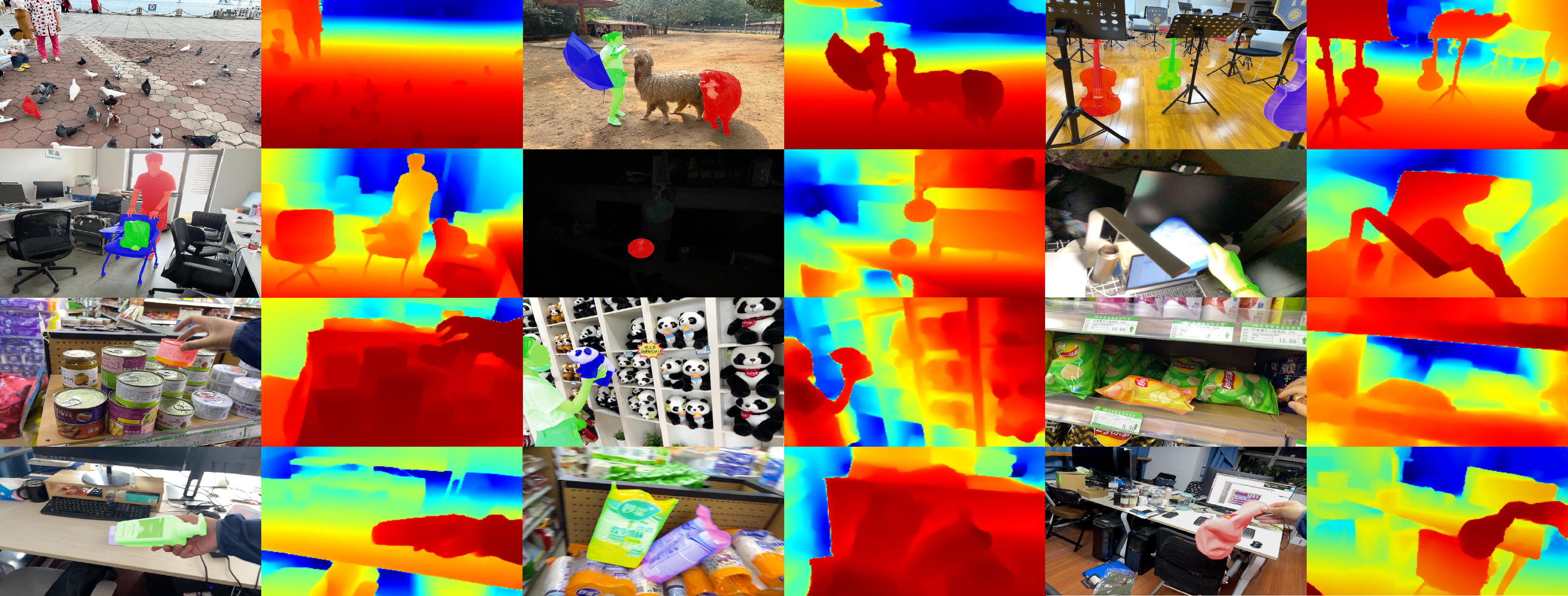}
    \caption{Samples from ARKitTrack. We capture both indoor and outdoor sequences (1st row) in many scenes, including zoo, market, office, square, corridor, etc. Lots of scenarios are presented in our dataset, e.g., low or high light conditions (2nd row), surrounding clutter (3rd row), out-of-plane rotation, motion blur, deformation, etc. (4th row). Besides, we annotate each frame with object masks.}
    \vspace{2mm}
\end{center}%
}]

\customfootnotetext{$\dag$}{Equal contribution}

\customfootnotetext{*}{Corresponding author: Dr. Lijun Wang, ljwang@dlut.edu.cn}

\begin{abstract}
    Compared with traditional RGB-only visual tracking, few datasets have been constructed for RGB-D tracking.
    In this paper, we propose ARKitTrack, a new RGB-D tracking dataset for both static and dynamic scenes captured by consumer-grade LiDAR scanners equipped on Apple's iPhone and iPad. 
    ARKitTrack contains 300 RGB-D sequences, 455 targets, and 229.7K video frames in total. Along with the bounding box annotations and frame-level attributes, we also annotate this dataset with 123.9K pixel-level target masks. Besides, the camera intrinsic and camera pose of each frame are provided for future developments.
    To demonstrate the potential usefulness of this dataset, we further present a unified baseline for both box-level and pixel-level tracking, which integrates RGB features with bird's-eye-view representations to better explore cross-modality 3D geometry. 
    In-depth empirical analysis has verified that the ARKitTrack dataset can significantly facilitate RGB-D tracking and that the proposed baseline method compares favorably against the state of the arts.
    The code and dataset is available at \url{https://arkittrack.github.io}.

\end{abstract}

\section{Introduction}
\label{sec:intro}
    %
    As a fundamental and longstanding problem in computer vision, visual tracking has been studied for decades and achieved significant progress in recent years with many advanced RGB trackers~\cite{SiamRPN2,ECO,MixFormer,SiamMask,2019FEELVOS,2021STCN} and datasets~\cite{LaSOT,GOT10k,TrackNet,YTBVOS} being developed. Nonetheless, there still exist many challenging situations such as occlusion, distraction, extreme illumination, etc., which have not been well addressed.
    %
    %
    With the wide application of commercially available RGB-D sensors, many recent works~\cite{tracker1,tracker3,tracker4,DAL,TSDM} have focused on the RGB-D tracking problem, as depth can provide additional 3D geometry cues for tracking in complicated environments. The development of RGB-D tracking is always boosted by the emergence of RGB-D tracking datasets. The early RGB-D datasets~\cite{STC,PTB} only have a limited number of video sequences and can hardly meet the requirement of sufficient training and evaluating sophisticated RGB-D trackers.
    To alleviate this issue, two larger datasets~\cite{CDTB,DepthTrack} have been built recently and successfully adopted in the VOT-RGBD challenges~\cite{VOT19,VOT20,VOT21}.

    Though existing RGB-D tracking datasets strongly benefit the development of RGB-D trackers, they are still limited in the following two aspects. First, these datasets are collected using Realsense or Kinect depth cameras, which require edge devices for onsite computing or post-processing and are not easily portable. As a result, it severely restricts the scene diversity, and most videos are captured under static scenes, which causes a large domain gap between the collected dataset and real-world applications. Second, existing RGB-D tracking datasets only contain bounding box-level annotations and mostly fail to provide pixel-level mask labels. Therefore, they are not applicable for training/evaluating pixel-level tracking tasks (i.e., VOS).


    With the recent launch of built-in depth sensors of mobile phones (e.g., LiDAR, ToF, and stereo cameras) and the release of AR frameworks (e.g., Apple's ARKit~\cite{ARKit} and Google's ARCore~\cite{ARCore}), it becomes more convenient than ever to capture depth data under diverse scenes using mobile phones. Compared to prior depth devices, mobile phones are highly portable and more widely used for daily video recording. Besides, the depth maps captured by consumer-grade sensors mounted on mobile phones are also different from previous datasets in terms of resolution, accuracy, etc. 
    
    In light of the above observations, we present ARKitTrack, a new RGB-D tracking dataset captured using iPhone built-in LiDAR with the ARKit framework. The dataset contains 300 RGB-D sequences, 229.7K video frames, and 455 targets. Precise box-level target locations, pixel-level target masks, and frame-level attributes are also provided for comprehensive model training and evaluation. 
    Compared to existing RGB-D tracking datasets, ARKitTrack enjoys the following two distinct advantages. First, ARKitTrack covers more diverse scenes captured under both static and dynamic viewpoints. Camera intrinsic and 6-DoF poses estimated using ARKit are also provided for more effective handling of dynamic scenes. Therefore, ARKitTrack is more coincide with real application scenarios, particularly for mobile phones. Second, to our best knowledge, ARKitTrack is one of the first RGB-D tracking datasets annotated with both box-level and pixel-level labels, which is able to benefit both VOT and VOS.     
    
    To demonstrate the strong potential of ARKitTrack, we design a general baseline RGB-D tracker, which effectively narrows the gap between visual object tracking (VOT) and segmentation (VOS). Most existing RGB-D trackers employ the low-level appearance cues (e.g., contours and regions) of depth maps but fail to explore the 3D geometry information. To remedy this drawback, we propose to integrate RGB features with bird's-eye-view (BEV) representations through a cross-view feature fusion scheme, where RGB feature is mainly used for target appearance modeling and BEV representations built from depth maps can better capture 3D scene geometry. Experiments on our ARKitTrack datasets demonstrate the merit of the baseline tracker.
    

    Our contribution can be summarized into three folds:
    \begin{itemize}
    \item A new RGB-D tracking dataset, ARKitTrack, containing diverse static and dynamic scenes with both box-level and pixel-level precise annotations. 
    \item A unified baseline method for RGB-D VOT and VOS, combining both RGB and 3D geometry for effective RGB-D tracking.
    \item In-depth evaluation and analysis of the new dataset and the baseline method, providing new knowledge to promote future study in RGB-D tracking. 
    \end{itemize}

\begin{table*}[t]
    \centering
    \caption{
        Overview of the existing RGB-D VOT datasets and RGB VOS datasets.
    }
    \vspace{-2mm}
    \resizebox{\linewidth}{!}{
            \begin{tabular}{c  c c c c c c c c c c c}
                \toprule[0.3mm]
                Dataset        & \makecell{Num.\\Tracks}& \makecell{Total\\Frames}  & \makecell{Max.\\Frames}  & \makecell{Avg.\\Frames}  & \makecell{Min.\\Frames}  & \makecell{Num.\\Mask} & \makecell{Camera\\Pose} & \makecell{VOS\\Subset} & \makecell{Train\\Subset} & \makecell{Num.\\Attr.} & Year\\
                \midrule[0.2mm]
                PTB~\cite{PTB}            & 100        & 21.5K         & - & 215           & -            & - & \XSolidBrush & \XSolidBrush & \XSolidBrush & 5 & 2013 \\
                STC~\cite{STC}            & 36         & 9.0K          & - & 250           & -             & - & \XSolidBrush & \XSolidBrush & \XSolidBrush & 12& 2018 \\
                CDTB~\cite{CDTB}           & 80         & 101.9K        & 2,501 & 1,274         & 406           & - & \XSolidBrush & \XSolidBrush & \XSolidBrush & 13& 2019 \\
                DepthTrack~\cite{DepthTrack}     & 200        & 294.5K        & 4,069 & 1,473         & 143           & - & \XSolidBrush & \XSolidBrush & \Checkmark & 15& 2021 \\
                \midrule[0.2mm]
                DAVIS~\cite{DAVIS}           & 193       & 6.2K          & 104   &69         & 25            & 13.5K     & \XSolidBrush & \Checkmark& \Checkmark  &0 & 2017 \\
                YoutubeVOS~\cite{YTBVOS}      & 6459      & 94.6K         & 36    &27         & 4             & 166.2K    & \XSolidBrush & \Checkmark & \Checkmark &0 & 2019 \\
                \midrule[0.2mm]
                ARKitTrack          & 455        & 229.7K        & 2,566 & 765           & 110            & 123.9K  & \Checkmark & \Checkmark & \Checkmark & 16& 2022  \\
                \bottomrule[0.3mm]
            \end{tabular}
    }
    \label{overall}
    \vspace{-2mm}
\end{table*}

\section{Related Work}
    \subsection{RGB-D Tracking Datasets}
        Four datasets have been proposed for the RGB-D general object tracking task.
        Princeton Tracking Benchmark (PTB)~\cite{PTB} is constructed to alleviate the lack of RGB-D tracking datasets. This dataset is captured with Kinect and contains $100$ indoor video sequences. Sequences that have full occlusion frames in this dataset can evaluate the re-detection ability, which is important for long-term tracking. However, this dataset has a calibration problem, approximately $14\%$ of sequences have channel synchronized error and approximately $8\%$ suffer from miss-alignment. Besides, only $5$ validation videos have accessible ground truth.
        The Spatio-Temporal Consistency (STC) dataset~\cite{STC} uses Asus Xtion to capture the RGB-D videos. To enrich the data diversity, this dataset captures both indoor and outdoor scenarios. Although there are only $36$ sequences, this dataset provides binary and statistical attribute annotations for each frame.
        Color-and-depth general visual object tracking benchmark (CDTB)~\cite{CDTB} is proposed for VOT-RGBD challenge~\cite{VOT19}. This dataset is constructed for more realistic settings. It has $80$ long-term video sequences and the average length is $ 1,274$, which is approximately six times longer than PTB and STC. In these sequences, the average length of target absence periods is nearly ten times greater than PTB. All these long-term properties make this dataset more challenging and realistic. To increase the data diversity, this dataset captures more dynamic scenarios and more target pose changes. Besides, various devices are used for data collection, including Kinect, ToF-RGB pair, and Stereo Camera.
        Recent DepthTrack~\cite{DepthTrack} uses Intel Realsense 415 to capture video sequences. This large-scale dataset contains $150$ training sequences and $50$ test sequences, with an average sequence length of $1,473$. To further increase data diversity, DepthTrack captures $40$ scene types, $90$ object types, and annotates each frame with $15$ attributes.

    \subsection{RGB-D Visual Object Tracking}
        The development of RGB-D trackers is always limited by the related datasets. By constructing the PTB dataset, Song et al.~\cite{PTB} propose two types of RGB-D trackers. The first tracker extracts both the RGB feature and depth feature. To exploit the depth information, this tracker extracts HOG~\cite{HOG} feature on the depth map and 3D features in point cloud space. Then a classifier is trained for detection-based tracking. The second tracker is based on the 3D point cloud and predicts the 3D bounding box.
        Zhong et al.~\cite{tracker1} propose to extract a set of 3D context key points on the dense depth map and use it for collaborative tracking in RGB-D videos.
        Bibi et al.~\cite{tracker2} build the appearance model and motion model in the 3D space. By using the particle filter framework, they propose a part-based sparse tracker.
        Kart et al.~\cite{tracker3} propose a generic RGB-D tracking framework and convert three Discriminative Correlation Filter (DCF) RGB trackers into RGB-D trackers. In this framework, the depth map is used for foreground segmentation, which can provide a strong cue for occlusion.
        DAL~\cite{DAL} embeds the depth information into deep features and learns a depth-aware discriminative correlation filter. It uses the depth map to generate a modulation map, which can re-weight the learned base filter to reduce the effect of the background and occlusion. Besides, the depth histogram is used for the target presence indicator. With this indicator and a re-detection scheme, DAL can perform long-term RGB-D tracking.
        SiamOC~\cite{SiamOC} converts SiameseRPN~\cite{SiamRPN1} into a RGB-D tracker. This tracker uses the depth map not only for occlusion estimation but also to correct the target location. According to the different distributions of the target's depth histogram, the tracker can estimate the occlusion state. Then, it uses the occlusion state and a Kalman filter to correct the target location.
        TSDM~\cite{TSDM} uses the depth map to generate a target mask to enhance the SiameseRPN++ tracker~\cite{SiamRPN2}. Then, depending to the depth value, it cuts the predicted box to give a more precise result. Similarly, Xiao et al.~\cite{tracker4} propose a box adaptive strategy according to the depth value.
        Instead of using the depth map outside the base tracker, recent DeT~\cite{DepthTrack} converts the depth map to a pseudo color image and uses an additional tracker branch to fuse the depth information for tracking.
        
        Most works simply extract the appearance feature from the depth map, and the important geometric cues are ignored. In this work, we explore the geometric information in the BEV space and fuse appearance and geometric features jointly.        

    \subsection{Semi-supervised Video Object Segmentation}
        Semi-supervised VOS methods aim to transfer the manually labeled frames to the following unlabeled video sequence. A number of matching-based methods\cite{2021STCN, 2021LCM, 2018VideoMatch, 2010AFB, 2020EGMN, 2019STM, 2020KMN, 2021HMMN, 2021SwiftNet, 2018SiamMask, 2019RANet, 2021RMNet} take Semi-VOS as a propagation task. They use the first or intermediate frames as templates to match with future frames. Therefore, how to better use spatial and temporal information in a sequence is always the mainstream research direction. FEELVOS\cite{2019FEELVOS} and CFBI\cite{2020CFBI} perform pixel-level matching globally and locally using both the first frame and previous adjacent frames. STM\cite{2019STM} and the following methods\cite{2021STCN, 2021LCM, 2020KMN, 2021HMMN, 2021RMNet} leverage a memory network to store intermediate frames as references and apply non-local attention mechanisms to match and propagate target embeddings. Based on the identification mechanism, AOT\cite{2021AOT} can embed multiple target objects into the same embedding space and then propagate all object embeddings uniformly and simultaneously. 
        
        All the existing algorithms are developed with only RGB datasets~\cite{YTBVOS,DAVIS}.
        In this work, we present the first large-scale RGB-D VOS dataset to unveil the potential of RGB-D segmentation. A novel method is also developed with this new dataset, showing satisfactory performance.

\section{The ARKitTrack Dataset}
    \subsection{Sequences Collection}
        Existing RGB-D tracking datasets are captured with depth sensors, such as Kinect and Realsense, which are linked with backend computing devices and not easily portable, resulting in the limited scene diversities. To build a more practical dataset, we collect RGB-D videos using a mobile phone, i.e., iPhone 13 Pro, which has a 12 MP wide-angle color camera and a LiDAR scanner. An iOS APP has also been developed to process and export the captured RGB-D videos at 30 FPS. Each video sequence contains synchronized and aligned RGB frames, depth maps, and confidence maps. 
        %
        The RGB frames are stored using $1920\times1440$ resolution in JPEG format under a low compression rate. The depth maps and confidence maps are processed with ARKit~\cite{ARKit} and stored using $256\times192$ resolution in 32-bit TIFF and PNG format, respectively. In confidence maps, $2$, $1$, and $0$ indicate high, medium, and low accuracy levels, respectively. To facilitate 3D and AR applications in dynamic scenes, we also provide the camera intrinsic and 6-DoF camera poses of each frame.

        During data collection, we ensure scene diversity from the following two aspects. To improve viewpoint diversity, we elaborately capture both static and dynamic scenes. Rich camera motion can cause complicated appearance and depth changes, delivering additional challenges to RGB-D tracking, which mimics real-world application scenarios.
        %
        To enrich the video content diversity, we capture both indoor videos and outdoor videos under different illumination conditions in a large number of different scenes, including zoo, market, office, street, square, corridor, etc. To further improve tracking difficulties, a large proportion of the sequences contain distracting objects with a similar appearance to the target. Besides the above features, the ARKitTrack dataset also contains many other challenging factors which have been classified into 16 attributes (See Sec.~\ref{sec:data_ann}).
        %

        The final ARKitTrack dataset contains $300$ sequences with $455$ targets, which includes 144 object categories and 287 dynamic scenes, exceeding~\cite{DepthTrack} (90/44) and~\cite{CDTB} (21/0). Table~\ref{overall} compares the overall properties between ARKitTrack and existing RGB-D VOT and VOS datasets. We select $50$ test sequences for VOT, which have rich scenes and motion patterns, and are sufficiently challenging for visual tracking. The average length of the VOT test set is $1,286$ frames and therefore can be used for long-term tracking evaluation.
        For VOS, we also select $55$ different test sequences at an average length of $328$ frames. These sequences have complicated contents and are longer than many previous VOS sequences, giving more challenges to segmentation algorithms.
        For each test set, the remaining video sequences are used as the training data (250 and 245 training videos for VOT and VOS, respectively).

        \begin{figure}[t]
            \centering
                \includegraphics[width=0.47\textwidth]{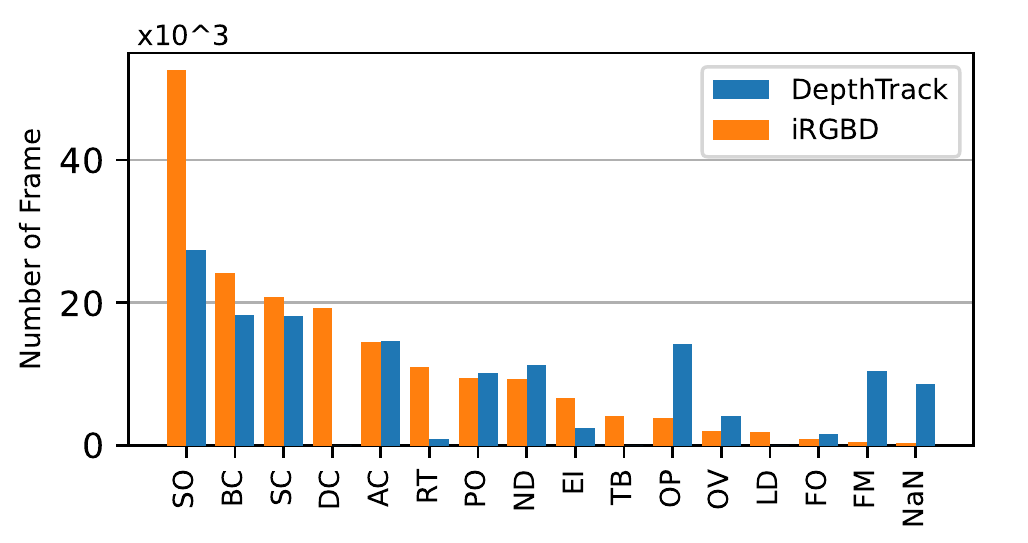}
            \vspace{-2mm}
            \caption{
                Attribute statistic comparison between ARKitTrack and DepthTrack test set.
            }
            \label{atts}
            \vspace{-8mm}
        \end{figure}

        \begin{figure*}[t]
            \centering
                \includegraphics[width=\textwidth]{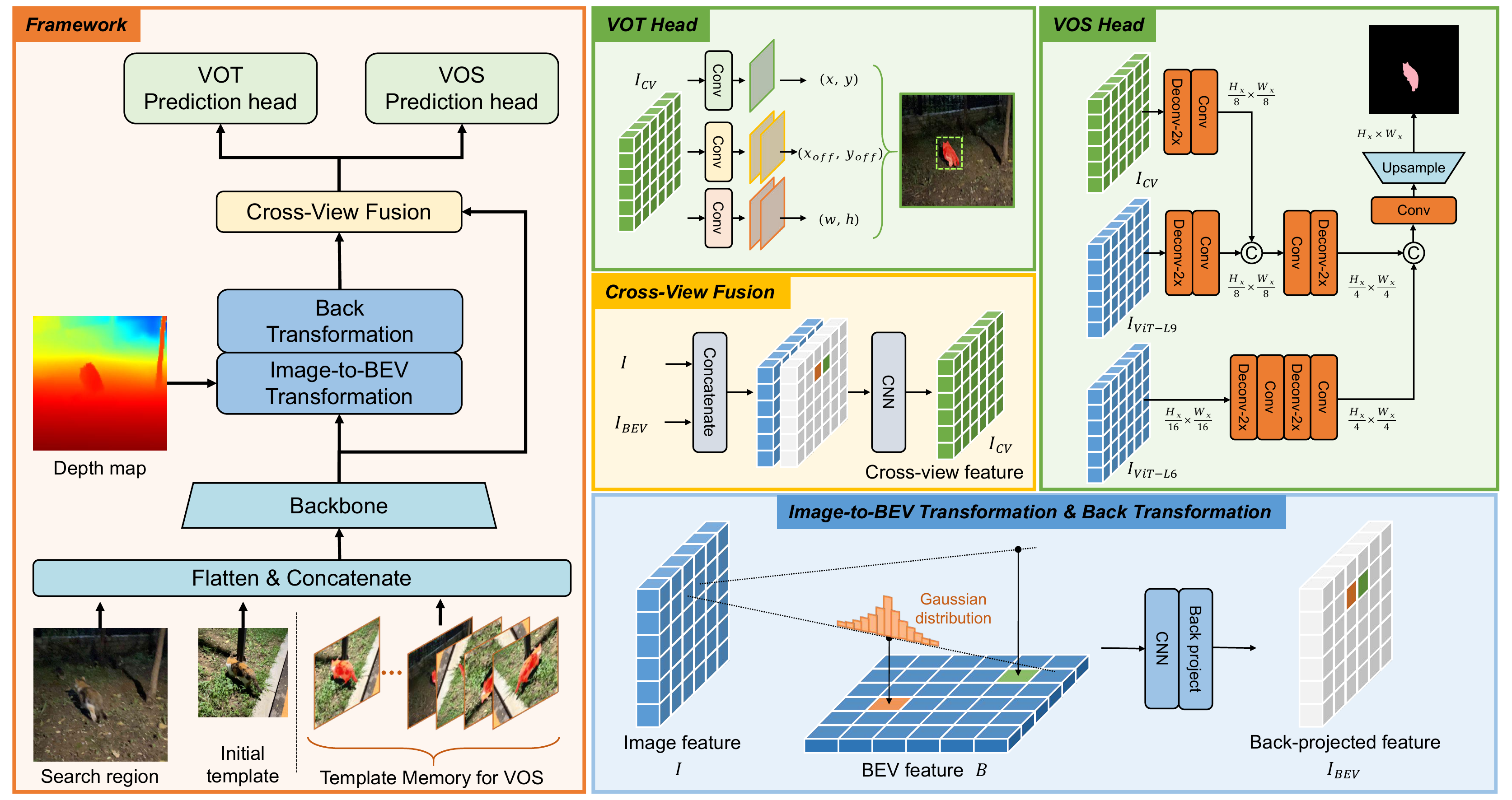}
            \caption{
                Overview the proposed unified RGB-D tracking pipeline in both box-level (VOT) and pixel-level (VOS).
            }
            \label{framework}
            \vspace{-3mm}
        \end{figure*}
        
    \subsection{Data Annotation}\label{sec:data_ann}
        The proposed dataset has three types of manual annotations, including axis-aligned target bounding boxes, pixel-level target masks, and per-frame attributes. 
        %
        Annotating object masks is always an expensive work. Most previous works (e.g.~\cite{YTBVOS,Pengyu}) annotate masks in a low frame rate or only annotate some keyframes.
        To annotate object masks in our dataset, we first ask annotators with expertise in VOT and VOS to manually select the appropriate sequences and target objects for annotation, and then uniformly sample key frames for every 3-4 frames, leading to the frame rates of 7-10 FPS. After that, we ask annotators to carefully annotate each keyframe with pixel-level target masks. The bounding box annotations for key frames can be automatically generated based on the target masks (i.e., axis-aligned boxes that most tightly enclose the target regions). 
        Box annotations of remaining frames are generated using the box interpolation method and manually refined by annotators.

        We also annotate $16$ per-frame challenging attributes for the VOT test sequences to enable attribute-level analysis. Among them, we borrow 12 attributes from prior works~\cite{DepthTrack,CDTB}, including Aspect-ratio Change (AC), Background Clutter (BC), Non-rigid Deformation (ND), Fast Motion (FM), Full Occlusion (FO), Out-of-plane Rotation (OP), Out-of-view (OV), Partial Occlusion (PO), Reflective Targets (RT), Size Change (SC), Similar Objects (SO), and Unassigned (NaN). 
        Another 4 attributes are specifically designed for our dataset, including Depth Clutter (DC), Extreme Illumination (EI), Low Depth Quality (LD), and Target Blur (TB). The LD label is calculated from the confidence map, which is predicted by ARKit and can provide useful information for realistic RGB-D applications. DC, EI, and TB reflect the diversity of scenes in our dataset, where rich motion scenarios cause TB frames. Please refer to the supplementary materials for details.
        %
        We use the VOT annotation tool~\cite{Aibu} to annotate frame-level attributes for the VOT test set. 
        Figure~\ref{atts} compares the attribute distributions of ARKitTrack and DepthTrack. 
        Our test sequences contain more distracting objects, background clutter, size change, extreme illumination, and reflective target frames. 

    \subsection{Performance metrics}
        For box-level tracking, we construct a test set (ARKitTrack-VOT) with $50$ long-term sequences. Therefore, we use the VOT long-term evaluation protocol~\cite{CDTB}. According to this protocol, a tracker is required to perform one-pass tracking on each sequence and predict bounding boxes and confidence scores.
        The overall evaluation uses three metrics: Precision (Pr.), Recall (Re.), and F-score~\cite{CDTB}. Tracking precision uses the intersection-over-union (IoU) between the predicted box and groundtruth to measure the target localization accuracy when the target is visible. Tracking recall can measure the accuracy of visible prediction.
        Specifically, by setting a confidence threshold $\tau$, all predictions $A(\tau)$ with confidence that higher than the threshold will be used for evaluation, resulting in the following measures
        \begin{align}
            Pr(\tau) &= \frac{1}{N_p} \sum_{t \in \{t:A_t(\tau) \neq \emptyset\}} IoU(A_t(\tau), G_t), \\
            Re(\tau) &= \frac{1}{N_g} \sum_{t \in \{t:G_t \neq \emptyset\}} IoU(A_t(\tau), G_t), 
        \end{align}
        where $G$ is the groundtruth, $N_p$ is the number of non-empty predictions, i.e., $A(\tau) \neq \emptyset$, $N_g$ is the number of visible frames, i.e., $G \neq \emptyset$. The primary metric F-score is defined as  
        \begin{align}
            F(\tau) = \frac{2Pr(\tau)Re(\tau)}{Pr(\tau)+Re(\tau)}.
        \end{align}
        For each tracker, by computing precision and recall over all possible thresholds, we can draw a precision-recall plot and find the highest F-score as the final measure score. 

        For pixel-level tracking, we select $55$ long-term sequences to construct a VOS test set (ARKitTrack-VOS). On this subset, we use three overall evaluation metrics, including region similarity ($\mathcal{J}$), contour accuracy ($\mathcal{F}$), and their average ($\mathcal{J\&F} = (\mathcal{J} + \mathcal{F}) / 2$).
        The region similarity is defined as the intersection-over-union between the estimated mask and groundtruth mask. The contour accuracy is the F-measure based on the contour-based precision and recall. We calculate the mean value $\mathcal{J_M}$ and $\mathcal{F_M}$ over all test sequences as the final scores.

        \begin{table*}[t]
            \centering
            \caption{
                Comparison results on the ARKitTrack test set and existing popular RGB-D benchmarks, including DepthTrack and CDTB. 
            }
            \resizebox{\linewidth}{!}{
                    \begin{tabular}{l |ccc |ccc |ccc | c c}
                        \toprule[0.3mm]
                        \multirow{2}{*}{Tracker} & \multicolumn{3}{c|}{ARKitTrack} & \multicolumn{3}{c|}{DepthTrack} & \multicolumn{3}{c}{CDTB} & \multicolumn{2}{|c}{Description}\\
                            & Pr. & Re. & F-score & Pr. & Re. & F-score & Pr. & Re. & F-score & Type & Year\\
                        \midrule[0.2mm]
                        Start-ST101~\cite{Stark}         & 0.407 & 0.381 & 0.393     & 0.503 & 0.468 & 0.485     & 0.657 & 0.669 & 0.663      & RGB  & 2022\\
                        OSTrack~\cite{OSTrack}             & 0.440 & 0.440 & 0.440     & 0.572 & 0.563 & 0.567     & 0.713 & 0.686 & 0.699      & RGB  & 2022\\
                        MixFormer1k~\cite{MixFormer}         & 0.449 & 0.421 & 0.434     & 0.490 & 0.454 & 0.471     & 0.692 & 0.664 & 0.678      & RGB  & 2022\\
                        ToMP101~\cite{ToMP}             & 0.449 & 0.433 & 0.441     & 0.515 & 0.495 & 0.505     & 0.670 & 0.683 & 0.676      & RGB  & 2022\\
                        \midrule[0.2mm]
                        TSDM~\cite{TSDM}                & 0.389 & 0.292 & 0.334     & 0.442 & 0.363 & 0.398     & 0.647 & 0.543 & 0.591      & RGBD & 2021\\
                        ATCAIS~\cite{VOT20}              & 0.389 & 0.343 & 0.364     & 0.473 & 0.402 & 0.435     & 0.709 & 0.696 & 0.702      & RGBD & 2020\\
                        DAL~\cite{DAL}                 & 0.446 & 0.329 & 0.378     & 0.512 & 0.369 & 0.429     & 0.620 & 0.560 & 0.589      & RGBD & 2020\\
                        TALGD~\cite{VOT21}               & 0.428 & 0.352 & 0.386     & 0.494 & 0.424 & 0.456     & 0.630 & 0.596 & 0.613      & RGBD & 2022\\
                        DeT~\cite{DepthTrack}                 & 0.428 & 0.405 & 0.416     & 0.560 & 0.506 & 0.532     & 0.674 & 0.642 & 0.657      & RGBD & 2021\\
                        STARK\_RGBD~\cite{VOT21}         & 0.469 & 0.426 & 0.446     & 0.570 & 0.558 & 0.564     & \textbf{0.743} & \textbf{0.769} & \textbf{0.755}      & RGBD & 2022\\
                        DDiMP~\cite{VOT20}               & \textbf{0.495} & 0.413 & 0.450     & 0.540 & 0.475 & 0.506     & 0.703 & 0.689 & 0.696      & RGBD & 2020\\
                        \midrule[0.2mm]
                        Ours                & 0.488 & \textbf{0.469} & \textbf{0.478}     & \textbf{0.617} & \textbf{0.607} & \textbf{0.612}     & 0.711 & 0.671 & 0.690      & RGBD & 2022\\
                        \bottomrule[0.3mm]
                    \end{tabular}
            }
            \label{vot_overall}
        \end{table*}

\section{A Unified RGB-D Tracking Baseline}
    We introduce a new RGB-D tracking baseline, which unifies both box-level and pixel-level target tracking. 
    As opposed to existing RGB-D tracking methods that mainly explore the appearance cues in depth maps, we model 3D scene geometry from the BEV view and perform cross-view fusion to integrate appearance and geometry representations for robust RGB-D tracking. As illustrated in Figure~\ref{framework}, the overall pipeline consists of an image-view encoder, BEV transformer, cross-view fusion module, and task-specific inference heads. Architecture designs are detailed as follows. 
    
    \subsection{Cross-view Fusion}

        \textbf{Image-View Encoding.}
        We first use the ViT model~\cite{OSTrack} to extract image feature map $\mathbf{I}$, which takes the template-search image pair as input and jointly extracts the image feature. 
        Specifically, the initial template $\mathbf{Z} \in \mathbb{R}^{C_z \times H_z \times W_z}$\footnote{The initial template is the RGB image ($C_z = 3$) and the concatenation of the RGB image with initial target mask ($C_z = 4$) for VOT and VOS, respectively.} 
        and current frame search region $\mathbf{X} \in \mathbb{R}^{3 \times H_x \times W_x}$ are first split into $P \times P$ patches independently. Then, these patches are flattened and linearly projected into $D$ dimension token embeddings $\mathbf{E}_z \in \mathbb{R}^{N_z \times D}$ and $\mathbf{E}_x \in \mathbb{R}^{N_x \times D}$. After adding position embeddings, all the token embeddings are concatenated and fed into the ViT model~\cite{ViT} for feature extraction by encoding the correlation between the template and search regions. The enhanced search region tokens are taken as the output and reshaped into a 2D feature map $\mathbf{I} \in \mathbb{R}^{C_I \times H_I \times W_I}$.
        

        \textbf{Image-to-BEV Transformation.}
        The 2D depth maps suffer from a geometric-lossy problem~\cite{bevfusion}, i.e., faraway points in 3D space might be projected into neighboring pixels in the 2D image plane. Encoding the depth map in the 2D form is therefore a sub-optimal way to exploit geometry information. Instead, we transform the depth map into the BEV space and modulate the BEV feature for better depth map encoding.
        Specifically, we take the encoded RGB feature map $\mathbf{I}$ as input and follow LSS~\cite{LSS} to embed both the RGB and depth information into a BEV feature map $\mathbf{B} \in \mathbb{R}^{C_B \times H_B \times W_B}$ with the pillar format~\cite{poingpillar} using an accelerated BEV pooling~\cite{bevfusion} and conv layers. However, different from LSS which predicts the discrete depth distribution, we model depth for each pixel using Gaussian distribution centered at the input depth value.

        The BEV space can better capture 3D geometry, where neighboring points are close to each in 3D space with similar depths. Therefore, we directly use a convolutional sub-network to further modulate the BEV feature, which can effectively aggregate information within a 3D local context and also compensates for the imperfect BEV transformation caused by inaccurate depth. 

        \textbf{Image-BEV Cross-View Fusion.}
        In order to perform cross-view fusion, the corresponding features under 2D image and BEV views should be first spatially aligned. Since target localization is conducted on the image plane, we back-project the BEV feature to the 2D image space, producing $\mathbf{I}_{BEV} \in \mathbb{R}^{C_B \times H_I \times W_I}$, as illustrated in Figure~\ref{framework}. For a pixel $(i,j)$ on the image plane, we first project it to 3D and compute its BEV space coordinates $(k,l,p)$ according to its depth $d(i,j)$ and camera intrinsic. We then sample the pillar features $B(k,l)$ from the BEV view using nearest neighbor interpolation for efficiency, though more sophisticated sampling techniques are also applicable. The above procedure is conducted for each pixel and the sampled BEV features can be assembled into the well-aligned image-view feature map $\mathbf{I}_{BEV}$. 
        
        Finally, The image feature $\mathbf{I}$ and BEV feature $\mathbf{I}_{BEV}$ are fused via concatenation and conv layers to produce the final feature map $\mathbf{I}_{CV}$ which is further used for downstream tracking tasks. 
      
    \subsection{Bounding Box and Pixel-level Tracking}
        Our baseline method provides a general framework for both box and pixel-level tracking tasks (i.e., VOT and VOS), where only task-specific heads with minimum architectural modification are required to build upon the fused features.
        
        \textbf{VOT Head.}
        The VOT head is a fully-convolutional network with three convolution layers, which consumes the cross-view feature $\mathbf{I}_{CV}$ to predict a score map $\mathbf{L} \in \mathbb{R}^{1 \times H_I \times W_I}$, an offset map $\mathbf{O} \in \mathbb{R}^{2 \times H_I \times W_I}$, and a box size map $\mathbf{S} \in \mathbb{R}^{2 \times H_I \times W_I}$. The target center coordinate $(x,y)$ is determined by the peak response at the score map. We then obtain the corresponding center offset $(x_{off}, y_{off})$ and box size $(w, h)$ from $\mathbf{O}$ and $\mathbf{S}$, respectively. The final target bounding box is located at
        \begin{equation}
            (x,y,w,h) = (x + x_{off}, y + y_{off}, w, h).
        \end{equation}

        \textbf{VOS Head.}
        To improve pixel-level segmentation accuracy, we follow prior methods~\cite{2019STM, 2021STCN} by using temporal dynamic templates and multi-scale features. Similar to~\cite{Stark,MixFormer}, we introduce an IoU prediction branch to control the update of dynamic templates. If the predicted IoU is higher than a threshold, we memorize the current frame and mask as a new dynamic template, and the initial template is concatenated with all the dynamic templates for subsequent tracking. As shown in Figure~\ref{framework}, the VOS head built on top of the multi-scale feature map predicts the final target segmentation map $\mathbf{M}$ with convolutional layers interleaved by deconvolution upsampling.

        \begin{table}[t]
            \caption{Comparison results on the ARKitTrack VOS test set with existing popular VOS methods.}
            \label{vos_overall}
            \centering
            \footnotesize\setlength{\tabcolsep}{4pt}
            \resizebox{0.48\textwidth}{!}{
            \begin{tabular}{c|cccc|cc}
                \toprule[0.3mm]
                Segmentor        &STCN~\cite{2021STCN}   &AOT~\cite{2021AOT}    &RPCM~\cite{2022RPCM}   &QDMN~\cite{2022QDMN}   &STCN\_RGBD & \textbf{Ours}\\
                \midrule[0.2mm]
                Year    & 2021  & 2021  & 2022  & 2022  & 2021      & 2022\\
                Type    & RGB   & RGB   & RGB   & RGB   & RGBD      & RGBD\\
                \midrule[0.2mm]
                $\mathcal{J\&F} \uparrow $              & 0.525 & 0.582 & 0.509 & 0.306   & 0.537   & \textbf{0.662} \\
                $ \mathcal{J}_\mathcal{M} \uparrow $    & 0.489 & 0.555 & 0.492 & 0.276   & 0.498   & \textbf{0.625} \\
                $ \mathcal{F}_\mathcal{M} \uparrow $    & 0.560 & 0.627 & 0.527 & 0.337   & 0.575   & \textbf{0.698} \\
                \bottomrule
            \end{tabular}
        }
        \end{table}

    \subsection{Training Loss}
        For the VOT head, we use the weighted focal loss for score map prediction and $\ell_1$ loss with generalized IoU loss for offset and box regression:
        \begin{equation}
        \label{equ:vot_loss}
            \begin{array}{cc}
                \mathcal{L}_{reg} = \lambda_{giou} \mathcal{L}_{giou} + \lambda_{\ell_1} \mathcal{L}_{\ell_1},\\
                \mathcal{L}_{track} = \mathcal{L}_{focal} + \mathcal{L}_{reg},
            \end{array}
        \end{equation}
        where $\lambda_{giou}=2$, $\lambda_{\ell_1}=5$ (please refer to~\cite{OSTrack} for details).
        
        For the VOS head, we use the dice loss~\cite{2016V-Net} and the bootstrapped cross entropy ~\cite{2021MiVOS} for masks prediction and the $\ell_2$ loss for IoU prediction:
        \begin{equation}
        \label{equ:vos_loss}
            \begin{array}{cc}
                {\mathcal{L}_{mask}} = {\lambda_{bce}}{\mathcal{L}_{bce}}{(\mathbf{M}, \mathbf{\hat{M}})} + {\lambda_{dice}}{\mathcal{L}_{dice}}{(\mathbf{M}, \mathbf{\hat{M}})}.\\
                {\mathcal{L}_{seg}} = {\mathcal{L}_{mask}} + \mathcal{L}_{\ell_2} + {\mathcal{L}_{reg}}{(\mathbf{b}, \mathbf{\hat{b}})}.
            \end{array}
        \end{equation}
        where $\mathbf{M}$ and $\mathbf{\hat{M}}$ are the predicted and groundtruth masks; $\mathbf{b}$ represents the bounding box enclosing the predicted target region and $\mathbf{\hat{b}}$ denotes the groundtruth bounding box. $\lambda_{bce} = 10$ and $\lambda_{dice} = 2$ are the regularization parameters.

        \begin{table}[t]
            \caption{The ablation analysis of our VOT method on the ARKitTrack test set and DepthTrack test set.}
            \label{vot_ab}
            \centering
            \resizebox{0.48\textwidth}{!}{
            \begin{tabular}{cccc|ccc|ccc|c|c}
                \toprule[0.3mm]
                \multicolumn{4}{c|}{Components}  & \multicolumn{3}{c|}{ARKitTrack} & \multicolumn{3}{c|}{DepthTrack~\cite{DepthTrack}}  & \multicolumn{1}{c|}{FLOPs}  & \multicolumn{1}{c}{Params} \\
                FT & BEV & CV & GD &Pr.    &Re.    & F-score &Pr.    &Re.    & F-score      & G & M\\
                \midrule[0.2mm]
                &&&                                         & 0.441 & 0.441 & 0.441     & 0.572 & 0.563 & 0.567 & 48.4  & 92.1\\
                \Checkmark&&&                               & 0.448 & 0.438 & 0.443     & 0.579 & 0.569 & 0.574 & 48.4  & 92.1\\
                \Checkmark&\Checkmark&&\Checkmark           & 0.427 & 0.420 & 0.423     & 0.570 & 0.537 & 0.553 & 56.2  & 104.1\\
                \Checkmark&\Checkmark&\Checkmark&           & 0.479 & 0.461 & 0.470     & 0.612 & 0.593 & 0.602 & 56.7  & 104.9\\
                \Checkmark&\Checkmark&\Checkmark&\Checkmark & \textbf{0.488} & \textbf{0.469} & \textbf{0.479}     & \textbf{0.617} & \textbf{0.607} & \textbf{0.612} & 56.7  & 104.9\\
                \bottomrule[0.3mm]
            \end{tabular}
            }
        \end{table}

\section{Experiments}
    We adopt pre-trained OSTrack384~\cite{OSTrack} to initialize the image-view encoder. All the other network parameters are randomly initialized. The VOT and VOS tasks are trained and evaluated separately by following existing experimental protocols~\cite{OSTrack,MixFormer,2019STM, 2021STCN} without otherwise stated. The trained model runs at 50FPS and 10FPS for VOT and VOS, respectively, at 384 input resolution on a single NVIDIA 3090 GPU. Detailed evaluation results are as follows.
    
    \subsection{Performance Evaluation}

        \textbf{Overall Performance of VOT.}
         We compare the proposed method with $7$ recent RGB-D trackers and $3$ state-of-the-art RGB trackers. The $7$ RGB-D tracker includes 3 recent methods (DeT~\cite{DepthTrack}, DAL~\cite{DAL}, and  TSDM~\cite{TSDM}) and the top four trackers from the VOT-RGBD 2021 challenge~\cite{VOT21} (STARK\_RGBD, TALGD, ATCAIS, DDiMP). Among them, ATCAIS and DDiMP are also the top trackers of VOT-RGBD 2020 challenge~\cite{VOT20}. The 4 state-of-the-art RGB trackers are ToMP~\cite{ToMP}, STARK~\cite{Stark},  MixFormer~\cite{MixFormer}, and OSTrack~\cite{OSTrack}.

        We perform evaluations on our ARKitTrack-VOT test set, DepthTrack test set~\cite{DepthTrack}, and CDTB~\cite{CDTB}, as shown in Table~\ref{vot_overall}. For all trackers, the overall performance on our dataset is always lower than that on DepthTrack and CDTB. It confirms that the proposed ARKitTrack is more challenging than the existing RGB-D tracking datasets.
        Our tracker performs better than other trackers on both ARKitTrack and DepthTrack, achieving $0.478$ and $0.612$ F-score, respectively. Although CDTB does not provide a training set and our tracker cannot gain from model learning on this dataset, our method still achieves satisfactory performance. $0.677$ F-score can be achieved by training with ARKitTrack, and $0.690$ F-score is achieved by training with DepthTrack.

        \textbf{Overall Performance of VOS.}
        Since there is no existing RGB-D VOS method, we select $4$ state-of-the-art RGB-VOS methods for comparison on the ARKitTrack-VOS test set, including STCN\cite{2021STCN}, RPCM\cite{2022RPCM}, AOT (SwinB-L)\cite{2021AOT} and QDMN\cite{2022QDMN}. Besides, We design a variant named STCN\_RGBD for RGB-D VOS by adding an additional depth branch to STCN and fusing RGBD features through concatenation. For a fair comparison, all methods are re-trained on ARKitTrack-VOS train set without static image pre-training.
        As shown in Table~\ref{vos_overall}, our method consistently outperforms all the other compared methods in terms of all metrics. Please refer to the supplementary materials for more comparisons.
        
\begin{figure}[t]
            \centering
                \includegraphics[width=0.5\textwidth]{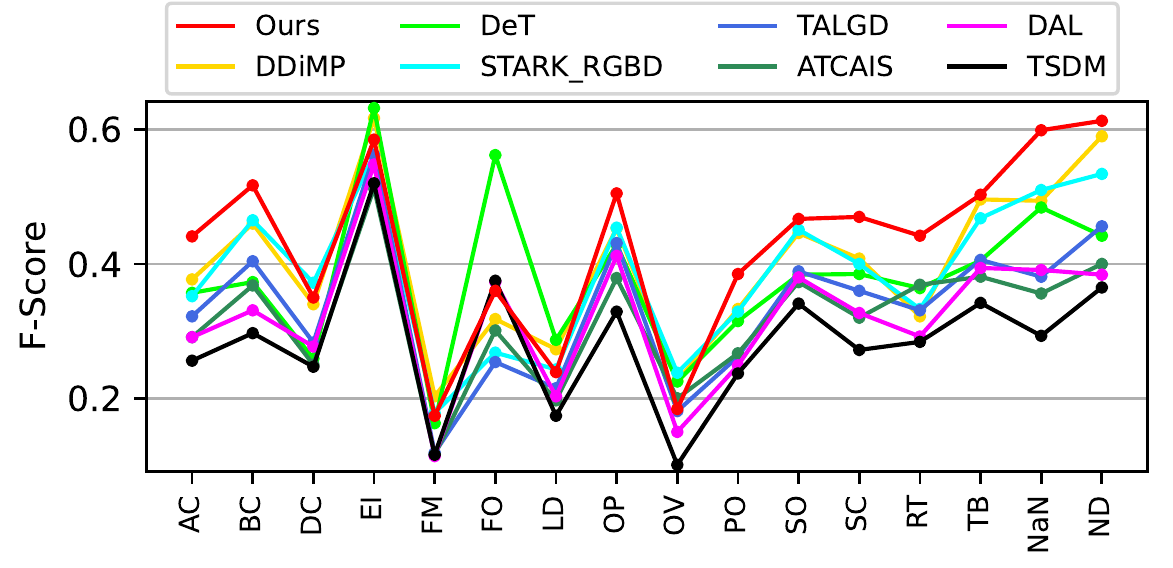}
            \caption{
                Attribute-specific analysis on the VOT test set.
            }
            \label{radar}
            \vspace{-2mm}
        \end{figure}

        \textbf{Attribute-based Performance.}
        We also conduct an attribute-specific analysis for the aforementioned RGB-D VOT trackers by using our per-frame attributes annotation, shown in Figure~\ref{radar}. It shows that our tracker performs better than other RGB-D trackers on 10 attributes. Besides, DeT outperforms other trackers on the full-occlusion attribute. 
        All trackers can deliver satisfactory performance on the extreme-illumination factor. However, none of them can well address the fast-motion and out-of-view attributes, indicating these attributes are more challenging for existing RGB and RGB-D trackers.

    \subsection{Ablation Study}
        \textbf{Studies on VOT.}
        To better demonstrate our method, we conduct ablation studies on both ARKitTrack and DepthTrack, shown in Table~\ref{vot_ab}. Our tracker is trained with their respective training sets and tested on their respective test sets. The basic tracker is OSTrack, which is trained with only the common RGB tracking datasets. By fine-tuning (\textbf{FT}) on the corresponding RGB-D training set with only RGB channels, this basic tracker achieves a slight improvement on both test sets. In this work, we propose embedding the depth channel into BEV space (\textbf{BEV}) and fusing RGB-D information from cross views (\textbf{CV}). By equipping the basic tracker with these methods, the tracker (\textbf{FT+BEV+CV}) gains $2.7\%$ and $2.8\%$ F-score on two test sets, respectively. After we apply the Gaussian distribution (\textbf{GD}) method in the BEV transformation, the final tracker (\textbf{FT+BEV+CV+GD}) further gains about $1\%$ F-score on both test sets. We also attempt to use only the back-projected BEV feature for tracking (\textbf{FT+BEV}), but the results are the worst. It shows that the back-projected BEV feature can enhance the image-view feature but is not equivalent to the image-view feature. It also suggests the effectiveness of our cross-view fusion module.

        \begin{table}[t]
            \caption{Comparison with different memorizing strategies.}
            \label{vos_dynamic}
            \centering
            \resizebox{0.42\textwidth}{!}{
            \begin{tabular}{ccc|ccc}
                \toprule[0.3mm]
                \multicolumn{3}{c|}{Components}  & \multicolumn{3}{c}{ARKitTrack}  \\
                IP & ONE & ADD &$\mathcal{J\&F} \uparrow$    &$\mathcal{J}_\mathcal{M} \uparrow$    & $\mathcal{F}_\mathcal{M} \uparrow$ \\
                \midrule[0.2mm]
                &&                                         & 0.630 & 0.594 & 0.666     \\
                &\Checkmark&                               & 0.642 & 0.604 & 0.680     \\
                \Checkmark&\Checkmark&                     & 0.648 & 0.611 & 0.685     \\
                \Checkmark&&\Checkmark                     & \textbf{0.662} & \textbf{0.625} & \textbf{0.698}     \\
                \bottomrule[0.3mm]
            \end{tabular}
            }
            \vspace{2mm}
        \end{table}

        \textbf{Studies on VOS.}
        We also explore different memory update strategies for VOS, and the results are shown in Table~\ref{vos_dynamic}.
        Our method is trained with the ARKitTrack-VOS training set and tested on the ARKitTrack-VOS test set. The baseline is our method that uses the designed VOS head with no dynamic templates achieving $0.630 \mathcal{J\&F}$. \textbf{ONE} indicates that we keep only one template and update it every N frames which improves $ \mathcal{J\&F}$ by $1.2\%$. By using the IoU prediction branch (\textbf{IP}) to control the template update, \textbf{IP+ONE} gains $1.8\% \mathcal{J\&F}$.
        \textbf{ADD} indicates that we consecutively add new templates to the memory bank and maintain several templates for prediction. By using this strategy, \textbf{IP+ADD} further gains $1.4\% \mathcal{J\&F}$ and achieves $0.662 \mathcal{J\&F}$. These results show the effectiveness of our VOS method and suggest the significance of the temporal information in the VOS task.

\section{Limitation and Ethical Concern}

    The proposed baseline tracker fails to exploit camera pose information. However, we believe that this information can provide essential cues for target tracking from dynamic viewpoints. Therefore, we will release these pose data along with the ARKitTrack dataset and explore this topic in future studies.
    We have obtained the approval for all the human targets in our dataset. For background pedestrians, we remove all of the personally identifiable information by blurring their faces for ethical concerns.

\section{Conclusion}

    We present ARKitTrack, a new large-scale RGB-D tracking dataset captured with iPhone built-in LiDAR under diverse scenes. Box annotations, pixel-level target labels, as well as frame-level attributes are provided to facilitate RGB-D VOT and VOS training\&evaluation. A general baseline tracker is also designed, which incorporates image appearance with BEV geometry through cross-view fusion and further closes up the gap between RGB-D VOT and VOS. We hope our dataset and baseline method can promote future research for the RGB-D tracking community. 

\vspace{2mm}
    \noindent\textbf{Acknowledgements.} This work is supported in part by National Natural Science Foundation of China (62293542, U1903215, 62276045), National Key R\&D Program of China (2018AAA0102001), Fundamental Research Funds for the Central Universities (DUT21RC(3)025, DUT22ZD210), and Dalian Science and Technology Talent Innovation Support Plan (2022RY17).

\clearpage
{\footnotesize
\bibliographystyle{ieee_fullname}
\bibliography{main}
}

\end{document}